\documentclass{article}
\usepackage{ijcai26}

\usepackage{times}
\usepackage{soul}
\usepackage{url}
\usepackage[hidelinks]{hyperref}
\usepackage[utf8]{inputenc}
\usepackage[small]{caption}
\usepackage{graphicx}
\usepackage{amsmath}
\usepackage{amsthm}
\usepackage{booktabs}
\usepackage{algorithm}
\usepackage{algorithmic}
\usepackage[switch]{lineno}

\usepackage{wrapfig}
\usepackage{amsfonts}
\usepackage{multirow}

\title{\emph{Context-Picker}: Dynamic Context Selection Using Multi-stage Reinforcement Learning}

\author{
Siyuan Zhu$^1$
\and
Chengdong Xu$^1$\and
Kaiqiang Ke$^1$\And
Chao Yu$^1$\\
\affiliations
$^1$Sun Yat-sen University\\
\emails
\{zhusy58, xuchd6, kekq\}@mail2.sysu.edu.cn,
yuchao3@mail.sysu.edu.cn,
}

\begin{document}

\maketitle

\begin{abstract}

In long-context question answering, selecting the appropriate scope of context for a query remains a key and unresolved challenge. Insufficient context can lead to missing essential information, whereas excessive context often introduces noise and degrades answer quality.
Conventional methods, such as retrieving a fixed number of passages or applying reranking,  struggle to dynamically determine which context to include. This is especially problematic for factoid questions, which typically depend only on a few precise pieces of evidence.
To overcome this limitation, we propose Context-Picker, a reasoning-aware framework that reframes context selection as the task of identifying a minimal sufficient evidence subset, moving beyond conventional similarity-based ranking. Context-Picker uses a human-inspired two-stage reinforcement learning schedule: stage 1 focuses on improving the recall rate of critical passages, and stage 2 prioritizes pruning redundancy to distill a compact evidence set.
To resolve reward sparsity, we propose an offline evidence distillation pipeline that mines ``minimal sufficient sets" via a Leave-One-Out (LOO) procedure, providing dense and task-aligned supervision.
Experiments on five long-context and multi-hop QA datasets demonstrate that our method outperforms strong RAG baselines and achieved higher answer accuracy.
Ablation studies also indicate that our coarse-to-fine optimization schedule, the redundancy-aware reward shaping, along with the rationale generated by the policy, all contribute substantially to these gains.

\end{abstract}

\section{Introduction}

\begin{figure}[t]
    \centering
    \includegraphics[width=\linewidth]{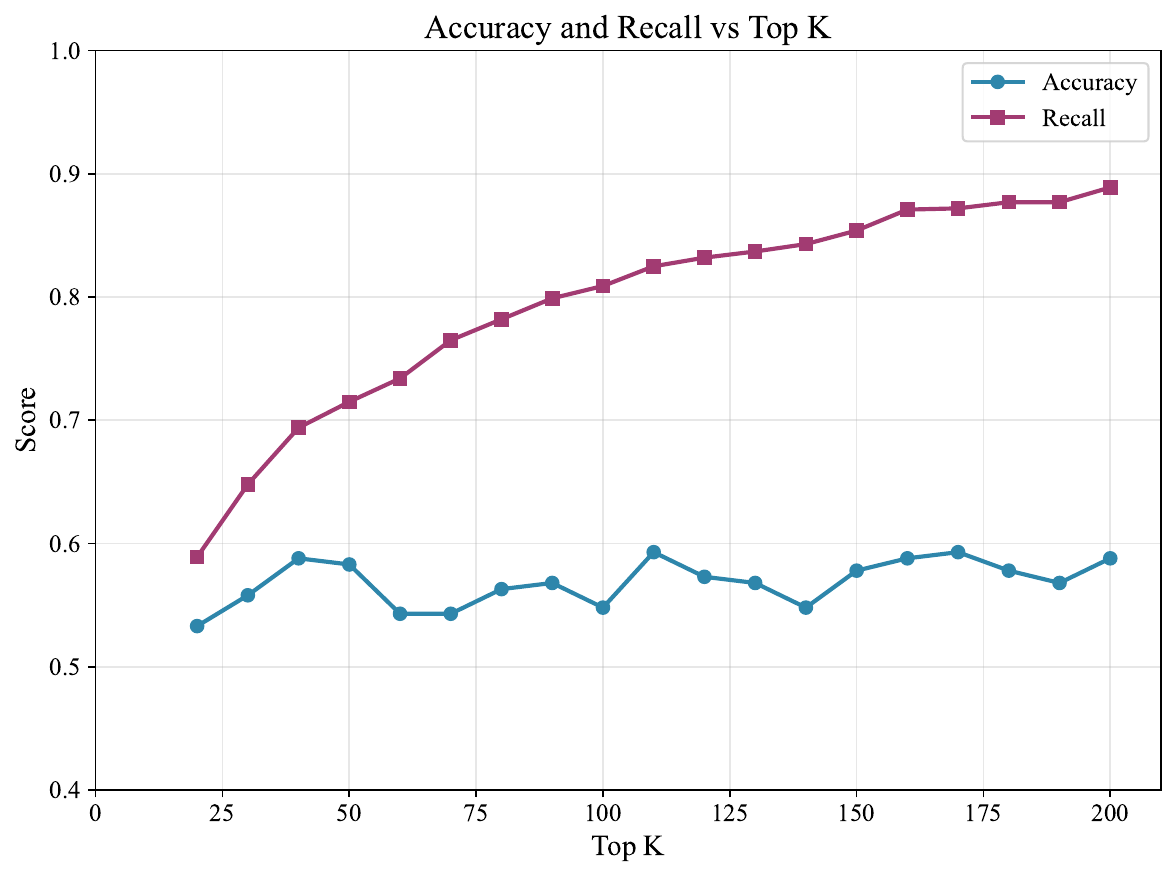}
    \caption{Accuracy versus retrieval depth (Top-$K$) in a standard RAG pipeline.
Recall increases with $K$, but answer accuracy does not improve, which is also reported in recent RAG studies~\protect\cite{jin2024longcontextllmsmeetrag}.}
    \label{fig:acc2topk}
\end{figure}

In long-context question answering (LCQA) tasks, Retrieval-Augmented Generation (RAG) has become a standard paradigm for Large Language Models (LLMs) to retrieve knowledge beyond their parametric knowledge~\cite{lewis2021retrievalaugmentedgenerationknowledgeintensivenlp,guu2020realmretrievalaugmentedlanguagemodel,izacard2022atlasfewshotlearningretrieval}. RAG enables LLMs to access query-related information and mitigates hallucination.
In practice, most RAG systems adopt a fixed Top-$K$ strategy:  a retriever ranks passages based on their semantic similarity with the query, and only the top-$K$ passages with highest similarity scores are concatenated and fed to the LLMs. Although this method is generally effective, it still faces the problem of identifying the suitable $K$ for each query.
If $K$ is too small, the model may miss critical evidence, leading to failure in answering the question. If $K$ is overly large, too many weakly related passages will result in degrading answer quality through distractors, attention dilution, and the "lost-in-the-middle" phenomenon, where LLMs underutilize information placed in the middle of long prompts~\cite{liu2023lostmiddlelanguagemodels}. As shown in Figure~\ref{fig:acc2topk}, increasing retrieval depth monotonically improves recall but often yields diminishing returns for answer accuracy~\cite{jin2024longcontextllmsmeetrag}, which suggests that context handling should be framed not as ranking, but as \emph{subset selection}--identifying a compact, query-specific evidence set sufficient for reasoning.

Recent studies on retrieval-augmented generation address this problem from two principal approaches. The first line of work enhances the retrieval pipeline while largely maintaining a fixed context size. Conventional sparse retrievers and dense dual-encoder models improve the recall using coarse ranking of candidate passages, and are frequently combined with cross-encoder or sequence-to-sequence rerankers that further refine the fine-grained ordering of documents~\cite{robertson2009probabilistic,karpukhin-etal-2020-dense,xiong2020approximatenearestneighbornegative,nogueira2020documentrankingpretrainedsequencetosequence}. More recent efforts employ LLM-based rerankers, which leverage query-aware, list-aware, or generator-aware signals to score and filter contexts~\cite{sun2024chatgptgoodsearchinvestigating,chen2025scirerankbenchbenchmarkingrerankersscientific,drozdov2023paradepassagerankingusing,wang2024learningretrieveincontextexamples,deng2025influenceguidedcontextselection}. While these methods are effective at elevating highly relevant passages and suppressing obvious distractors in the ranked list, the generator typically still processes either a fixed top-K set or manually defined thresholds; thus, the inherent trade-off between missing pertinent evidence and accumulating irrelevant noise still exists.

A complementary line of work \emph{adapts how much external evidence to use} on a per-query basis. Adaptive-RAG trains a lightweight classifier to route each query to different retrieval \emph{depths} (e.g., no retrieval, single-step retrieval, or iterative retrieval) based on estimated question complexity~\cite{jeong2024adaptiveraglearningadaptretrievalaugmented}. In contrast, Adaptive-$k$ selects a query- and context-specific cutoff $K$ by thresholding the similarity-score distribution of retrieved candidates, requiring neither additional fine-tuning nor extra LLM calls~\cite{taguchi2025efficientcontextselectionlongcontext}. While these approaches avoid using a fixed $k$, they rely on either a learned \emph{query-level} router or on \emph{non-learned} score-thresholding rules, and they do not explicitly optimize for identifying a non-redundant evidence subset that preserves answerability under a fixed input budget.

To move beyond routing and thresholding rules, reinforcement learning (RL) has recently been explored as a way to optimize retrieval and selection policies directly from task feedback while keeping test-time inference to a single policy forward pass. DynamicRAG models the reranker as an RL agent over document sequences and uses LLM-judged answer quality as reward to jointly adjust both the order and the number of retrieved documents~\cite{sun2025dynamicragleveragingoutputslarge}. Beyond reranking, recent RL-based systems such as Memory-R1 and related memory agents frame long-term memory management and retrieval decisions as RL problems, where a policy is trained to decide what to store, update, or retrieve in order to support downstream QA and dialogue~\cite{yan2025memoryr1enhancinglargelanguage}. RL has also been applied to conversational query reformulation, retrieval alignment and agentic RAG frameworks that optimize multi-step retrieval and reasoning trajectories~\cite{zhu2025convsearchr1enhancingqueryreformulation,xiong2025raggymsystematicoptimizationlanguage,jiang2025rexragreasoningexplorationpolicy}. However, many existing RL-style approaches still rely on \emph{trajectory-level} and relatively sparse rewards, making it difficult to credit assignment to individual passages and suppress redundant content. Moreover, they are often optimized for ranking quality, memory operations, or trajectory success rather than for explicitly finding the smallest evidence set that maintains answerability under a strict token budget.

In this paper, we introduce \emph{Context-Picker}, a reasoning-aware framework that fundamentally shifts the context selection paradigm from similarity-based ranking to minimal sufficient subset selection. Instead of treating retrieval as a sorting problem, Context-Picker formulates it as a decision-making process, and learns to construct a variable-length evidence set that is strictly necessary for answering the query.
Central to our approach is a human-inspired Coarse-to-Fine optimization strategy implemented via a two-stage reinforcement learning schedule.
In Stage I (Recall-Oriented), the picker is trained to maximize information with a relaxed redundancy margin, so that all potentially relevant passages—especially those spanning multiple passages—are captured.
In Stage II (Precision-Oriented), the objective shifts to refinement: the policy learns to prune redundant or weakly relevant passages, distilling the context into a compact, noise-free subset without compromising answerability.
To alleviate reward sparsity and stabilize training, we introduce an offline evidence distillation pipeline that uses a generator-judge loop with greedy Leave-One-Out (LOO) pruning to mine minimal sufficient evidence sets from raw documents. These distilled sets provide dense, task-aligned supervision, enabling the picker (i.e., policy) to learn to measure the contribution of each evidence piece.
Experiments on five long-context and multi-hop QA datasets demonstrate that our method outperforms strong RAG baselines and achieved higher answer accuracy.

\paragraph{Contributions.}
Our main contributions are three-fold: (1) We propose \emph{Context-Picker}, a reasoning-aware context picker trained with a two-stage reinforcement learning scheme (Recall-Oriented and Precision-Oriented) to jointly learn \emph{which} and \emph{how many} passages to pick. (2) We introduce an offline evidence mining pipeline that uses a generator-judge loop with Leave-One-Out (LOO) pruning to create minimal sufficient evidence sets for dense supervision. (3) We conduct extensive experiments on five long-context and multi-hop QA datasets, demonstrating that Context-Picker improves accuracy over strong RAG baselines on four out of five datasets.

\begin{figure*}[t]
    \centering
    \includegraphics[width=1.0\linewidth]{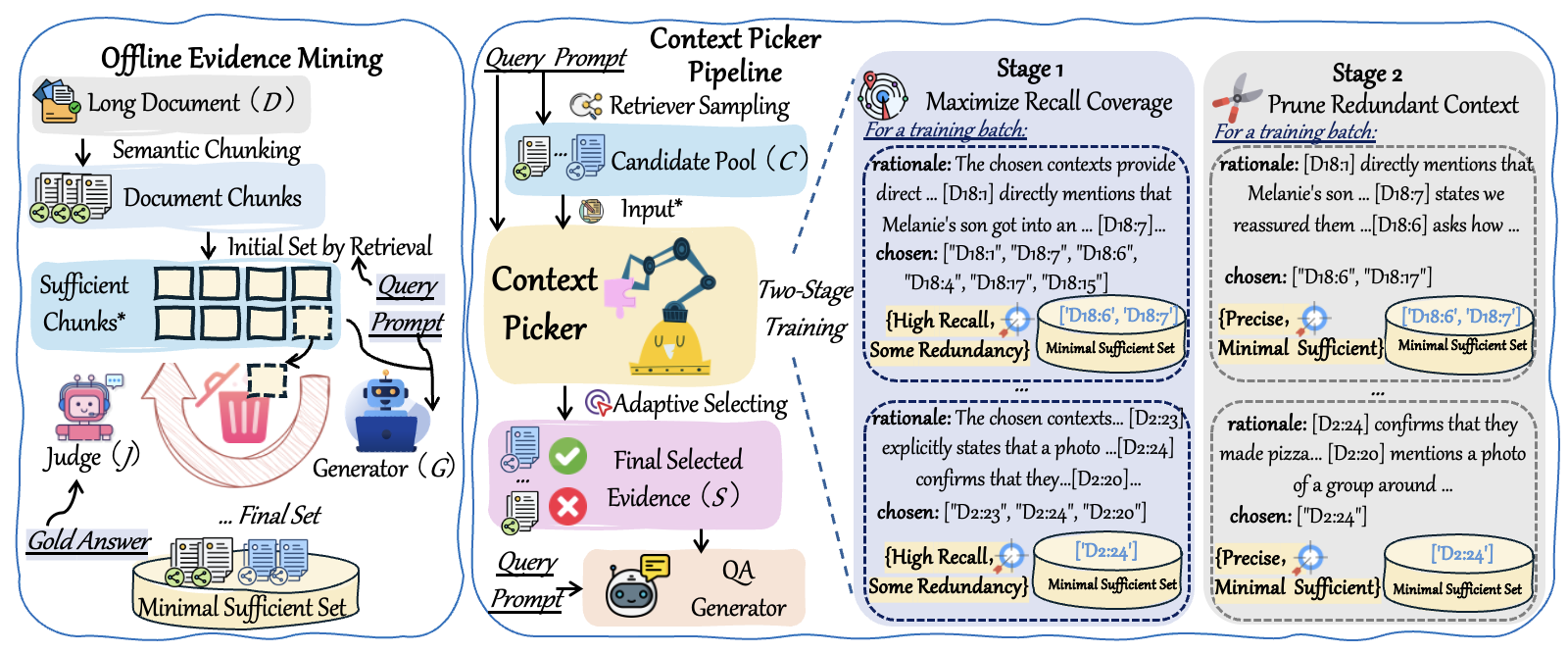}
    \caption{
        Overview of the \emph{Context-Picker} framework. The pipeline consists of two parts: 
        (1) \textbf{Offline Evidence Mining}, where a generator--judge loop employs a Leave-One-Out (LOO) strategy to mine minimal sufficient evidence sets ($\mathcal{S}_{\text{gold}}$) as supervision; 
        (2) \textbf{Context-Picker Pipeline}, where the picker policy ($\pi_\theta$) learns to select evidence from retrieved candidates ($\mathcal{C}$). The training follows a \textbf{Coarse-to-Fine} schedule: Stage I optimizes for high recall to capture reasoning chains, while Stage II tightens the redundancy penalty to distill a compact support set, guided by GRPO updates.
    }
    \label{fig:overview}
\end{figure*}

\section{Preliminaries}

\subsection{Retrieval-Augmented Generation}

We follow the standard retrieval-augmented generation (RAG) formulation
\cite{lewis2021retrievalaugmentedgenerationknowledgeintensivenlp,guu2020realmretrievalaugmentedlanguagemodel,izacard2022atlasfewshotlearningretrieval}.
Let $\mathcal{D}$ denote a large non-parametric corpus (e.g., Wikipedia or a long-term memory store).
In long-context QA, each document in $\mathcal{D}$ is first segmented into shorter passages (``chunks''), which serve as the retrieval units.
Given a query $q$, the retriever returns the top-$K$ most relevant passages.
\begin{equation}
\mathcal{C}(q) \;=\; \{ c_1, c_2, \ldots, c_N \}, \quad N \le K_{\max},
\label{eq:retrieval}
\end{equation}
optionally refined by a reranker that reorders $\mathcal{C}(q)$ according to query-specific relevance
\cite{karpukhin-etal-2020-dense,xiong2020approximatenearestneighbornegative,nogueira2020documentrankingpretrainedsequencetosequence}.
We use $\mathcal{C}(q)$, or simply $\mathcal{C}$ when the query is clear from context, to denote this (re)ranked candidate pool.
Later, when formulating Context-Picker, we additionally attach a unique identifier to each passage $c_j$ and write $\mathcal{C} = \{(c_j,\text{id}_j)\}_{j=1}^N$ for convenience.

\paragraph{Context selection.}
Given $\mathcal{C}(q)$, the system must choose a variable-length \emph{support set}
$\mathcal{S} \subseteq \mathcal{C}(q)$ to feed into the generator under an input budget $B$.
We view this as a subset selection problem that trades off task utility and brevity:

\begin{equation}
\begin{aligned}
\mathcal{S}^\star
&\in
\arg\max_{\mathcal{S} \subseteq \mathcal{C}(q)}
\Bigl(
  U(q, \mathcal{S}) - \lambda \cdot \mathrm{Len}(\mathcal{S})
\Bigr) \\
&\quad \text{s.t.} \quad
\mathrm{Tok}(q, \mathcal{S}) \le B .
\end{aligned}
\label{eq:selection_objective}
\end{equation}
where $U(q, \mathcal{S})$ is a task utility,
$\mathrm{Len}(\mathcal{S})$ measures the size of the support set,
$\lambda \ge 0$ controls the quality–brevity trade-off,
and $\mathrm{Tok}(q, \mathcal{S})$ counts input tokens.
Context-Picker builds on this formulation and learns a reasoning-aware policy under a token budget.

\paragraph{Response generation and utility.}
Given a support set $\mathcal{S}$, we construct a prompt
$x = \mathrm{Tpl}(q,\mathcal{S})$ by concatenating instructions, the query, and the selected passages,
and use a generator $\mathcal{G}$ to define a conditional distribution over answers:
$
p_\theta(y \mid x) \;=\; \mathcal{G}(x),
$
from which we decode an answer $\hat{y}$.
We instantiate the utility $U(q,\mathcal{S})$ in Eq.~\eqref{eq:selection_objective}
either with exact-match accuracy or with an LLM-as-judge score that evaluates the semantic correctness of $\hat{y}$ w.r.t.\ the reference answer.

\subsection{Optimization Objective}
We employ Group Relative Policy Optimization (GRPO)~\cite{shao2024deepseekmathpushinglimitsmathematical} to train the picker. Unlike PPO, GRPO eliminates the critic model by estimating baselines from group averages. For each query $q$, GRPO samples a group of outputs $\{a_i\}_{i=1}^G$ from the old policy $\pi_{\theta_{\mathrm{old}}}$:

\begin{equation}
\tiny
J(\theta) = \mathbb{E} \left[ \frac{1}{G} \sum_{i=1}^G \min \left( r_i(\theta) \hat{A}_i, \text{clip}(r_i(\theta), 1-\epsilon, 1+\epsilon) \hat{A}_i \right) \right] - \beta D_{KL}
\label{eq:grpo-objective},
\end{equation}
where $r_i(\theta) = \pi_\theta(a_i|q)/\pi_{\theta_{\mathrm{old}}}(a_i|q)$ and the advantage $\hat{A}_i$ is the normalized reward within the group.

\section{Context-Picker}

In this section, we introduce \emph{Context-Picker}, a reinforcement learning based method for selecting context passages. Figure~\ref{fig:overview} provides an overview of the proposed framework.
We start by formulating context picking as a single-step Markov decision process (MDP) in Section~\ref{subsec:rl-formulation}. Building on this formulation, we detail our overall framework, which comprises two key components: (i) an \emph{offline evidence mining} pipeline (Section~\ref{subsec:offline-mining}) that extracts compact yet sufficient evidence sets from raw documents, and (ii) a \emph{multi-stage reinforcement learning} scheme (Section~\ref{subsec:multi-stage-rl}) that learns a picker policy via a recall-oriented stage followed by a precision-oriented stage.
Finally, we detail how the learned picker is integrated with the downstream generator at inference time.

\subsection{Problem Formulation}
\label{subsec:rl-formulation}

We cast context picking as a single-step decision problem. Given a query $q$ and a stage-specific prompt $p_i$, the retriever provides a candidate pool $\mathcal{C} = \{(c_j, \text{id}_j)\}_{j=1}^N$. The observation is $o = \langle p_i, q, \mathcal{C} \rangle$. The policy outputs $\langle r, a \rangle$, where $r$ is a natural-language rationale and $a \subseteq \{\text{id}_j\}$ is a subset of selected IDs. This maps to the support set $\mathcal{S} = \{ c_j \mid (c_j, \text{id}_j) \in \mathcal{C}, \text{id}_j \in a \}$. Malformed or out-of-range selections are treated as invalid and penalized.

Given an offline-mined golden set $\mathcal{S}_{\text{gold}}$, the stage-$i$ reward is defined abstractly as:

\begin{equation}
R_i(o, a) = \mathrm{Cov}(\mathcal{S}, \mathcal{S}_{\text{gold}}) - \mathrm{Redun}_i(\mathcal{S}, \mathcal{S}_{\text{gold}}) - \gamma \,\mathbb{I}[\neg \text{valid}],
\label{eq:abstract-reward}
\end{equation}
where $\mathrm{Cov}$ measures coverage, $\mathrm{Redun}_i$ is a stage-dependent redundancy penalty, and the last term penalizes format errors. We instantiate this in Eq.~\eqref{eq:reward}.

\subsection{Data Curation}
\label{subsec:offline-mining}

\paragraph{Offline evidence mining.}
To construct high-quality training data for Context-Picker, we introduce an offline evidence distillation pipeline.
Each document $D$ is first segmented into semantically coherent chunks via semantic chunking, which ensures that each chunk forms a locally consistent unit while preserving contextual continuity.
This corresponds to the \emph{Offline Evidence Mining} module on the left side of Figure~\ref{fig:overview}, and the overall procedure is summarized in Algorithm~\ref{alg:test-time-evidence-picking}.

For each query–answer pair $(q, a)$, we perform retrieval using BM25 on the concatenation of the query and answer, i.e., on $[q; a]$ over the chunked document.
The top-$k$ retrieved chunks constitute an initial candidate set $\mathcal{S}_{\text{cand}}$.
We then run an answer-judge pipeline on $\mathcal{S}_{\text{cand}}$: a generator $\mathcal{G}$ produces a response $\hat{a}$ conditioned on $(q, \mathcal{S}_{\text{cand}})$, and an LLM-based judge $\mathcal{J}$ decides whether $\hat{a}$ semantically matches the gold answer $a$.
If $\mathcal{J}$ deems $\mathcal{S}_{\text{cand}}$ insufficient (i.e., the answer is judged as incorrect), we discard this pair, since the retrieved evidence does not support a correct answer even before pruning.

For the remaining pairs, we greedily prune redundant chunks via a leave-one-out (LOO) procedure.
We initialize $\mathcal{S}_{\text{suf}} \leftarrow \mathcal{S}_{\text{cand}}$ and iterate over chunks $c \in \mathcal{S}_{\text{suf}}$.
For each $c$, we temporarily remove it to form $\mathcal{S}' = \mathcal{S}_{\text{suf}} \setminus \{c\}$, run the same answer-judge pipeline on $(q, \mathcal{S}')$, and obtain a new judge decision.
If $\mathcal{J}$ still marks the answer as correct, we treat $c$ as redundant and permanently drop it, updating $\mathcal{S}_{\text{suf}} \leftarrow \mathcal{S}'$.
We repeat this LOO pruning until no chunk can be removed without flipping the judge decision from correct to incorrect.
The resulting set $\mathcal{S}_{\text{suf}}$ is thus a greedily minimal sufficient evidence set with respect to the judge: every remaining chunk is empirically necessary in the sense that removing any of them would cause the model to fail the judge.
We treat $\mathcal{S}_{\text{suf}}$ as the golden evidence supervision for training Context-Picker.

\paragraph{Data augmentation.}
Considered that most long-context QA or retrieval datasets contain relatively few unique queries samples, we introduce lightweight query rewriting to enhance data diversity. For each original query $q$, we generate five semantically equivalent but lexically diverse reformulations $\{q_i'\}_{i=1}^{5}$ using a language model. These rewrites preserve the meaning of the original query while varying in phrasing and focus, which helps improve linguistic diversity and reduces overfitting during RL training. During data partitioning, all rewrites of the same query are assigned to the same subset to prevent data leakage between training and evaluation data.

This curated dataset, consisting of golden evidence picks and diverse query formulations, serves as the foundation for the reinforcement learning phase of Context-Picker.

\begin{algorithm}[t]
\caption{Offline Evidence Mining}
\label{alg:test-time-evidence-picking}
\small
\begin{algorithmic}[1]
\REQUIRE Document $D$; query $q$; gold answer $a$;
         retriever $\mathcal{R}$; encoder $f_{\text{emb}}$;
         generator $\mathcal{G}$; answer judge $\mathcal{J}$;
         top-$k$
\ENSURE Minimal sufficient set $\mathcal{S}_{\text{suf}}$

\STATE $\mathcal{C} \leftarrow \text{Chunk}(D; f_{\text{emb}})$
      \hfill {\footnotesize // semantic chunking}
\STATE $x \leftarrow [q; a]$
\STATE $\mathcal{S}_{\text{cand}} \leftarrow \text{RetrieveTopK}(x, \mathcal{C}; \mathcal{R}, k)$
      \hfill {\footnotesize // initial evidence pool}
\STATE $\hat{a} \leftarrow \mathcal{G}(q, \mathcal{S}_{\text{cand}})$
\STATE $r_{\text{full}} \leftarrow \mathcal{J}(q, \hat{a}, a)$

\IF{$r_{\text{full}} = 0$}
    \RETURN $\emptyset$ \hfill {\footnotesize // if evidence insufficient, drop}
\ENDIF

\STATE $\mathcal{S}_{\text{suf}} \leftarrow \mathcal{S}_{\text{cand}}$
      \hfill {\footnotesize // start from full candidate set}
\STATE $\text{changed} \leftarrow \text{True}$

\WHILE{$\text{changed}$}
    \STATE $\text{changed} \leftarrow \text{False}$
    \FOR{each $c \in \mathcal{S}_{\text{suf}}$}
        \STATE $\mathcal{S}' \leftarrow \mathcal{S}_{\text{suf}} \setminus \{c\}$
              \hfill {\footnotesize // leave-one-out}
        \STATE $\hat{a}' \leftarrow \mathcal{G}(q, \mathcal{S}')$
        \STATE $r' \leftarrow \mathcal{J}(q, \hat{a}', a)$
        \IF{$r' = 1$}
            \STATE $\mathcal{S}_{\text{suf}} \leftarrow \mathcal{S}'$
                  \hfill {\footnotesize // if $c$ is redundant, prune}
            \STATE $\text{changed} \leftarrow \text{True}$
        \ENDIF
    \ENDFOR
\ENDWHILE

\RETURN $\mathcal{S}_{\text{suf}}$
\end{algorithmic}
\end{algorithm}

\subsection{Multi-stage Reinforcement Learning}
\label{subsec:multi-stage-rl}

Given the curated training set from Section~\ref{subsec:offline-mining}, this
subsection describes the two-stage policy optimization component of Context-Picker,
which corresponds to the right part of Figure~\ref{fig:overview}. We detail the two-stage optimization process below.

In long-context question answering (LCQA), the challenges of \emph{what evidence to pick} and \emph{how much evidence to pick} are essentially a coupled combinatorial optimization problem. 
As is shown in Figure \ref{fig:overview}, we decouple this problem into two training stages: \emph{a recall-oriented stage} that emphasizes comprehensive evidence coverage, and \emph{a refinement-oriented stage} that focuses on minimal sufficient selection.

\paragraph{Stage I: Recall-oriented strategy optimization.}
Stage~I is designed to learn a \emph{high-recall} picking behavior that prioritizes \emph{information completeness}.
In our setting, the downstream generator can answer a query correctly as long as the selected context set contains the key evidence that supports the reasoning chain.
We formalize this via the offline-mined minimal sufficient evidence set $\mathcal{S}_{\text{gold}}$ (Section~\ref{subsec:offline-mining}), which approximates the smallest subset that preserves answerability under an LLM-based judge.

Stage~I thus encourages the policy to maximize $\mathrm{Cov}(\mathcal{S}, \mathcal{S}_{\text{gold}})$ with a \emph{relaxed} redundancy tolerance $\mathrm{red}_1$ (Eq.~\ref{eq:reward}), allowing moderate over-selection.
This is crucial for multi-hop QA: missing even a single hop in the evidence chain can cause a failure, whereas including a few extra passages is often harmless at this stage.
By emphasizing coverage and using a loose redundancy margin, Stage~I prevents premature pruning and improves exploration over the combinatorial subset space, yielding a robust high-recall initialization for later compression.

\paragraph{Stage II: Refinement-oriented strategy optimization.}
Stage~II targets \emph{input conciseness} while preserving sufficiency, i.e., converging to a \emph{minimal sufficient evidence set}.
Starting from the high-recall policy learned in Stage~I, we tighten the redundancy margin to $\mathrm{red}_2<\mathrm{red}_1$ and strengthen the redundancy penalty in the reward (Eq.~\ref{eq:reward}),
so the policy is explicitly discouraged from keeping passages that do not improve answerability.
Intuitively, Stage~II pushes the picker toward solving a constrained compression problem:

\begin{equation}
\min_{\mathcal{S}\subseteq\mathcal{C}(q)} |\mathcal{S}|
\quad \text{s.t.}\quad
U(q,\mathcal{S})=1,
\label{eq:compression_view}
\end{equation}

where $U(q,\mathcal{S})$ is approximated during training by the distilled supervision $\mathcal{S}_{\text{gold}}$ and instantiated as coverage-plus-redundancy shaping in Eq.~\eqref{eq:reward}.
Operationally, Stage~II encourages the policy to keep sets that (i) retain near-complete coverage of $\mathcal{S}_{\text{gold}}$ (high recall), yet (ii) eliminate redundant, repetitive, or weakly relevant passages so as to reduce distractors and mitigate long-context degradation.
As a result, the learned picker progressively shifts from a \emph{recall-sufficient} regime to a \emph{precision-sufficient} regime, producing compact evidence subsets that maximize informativeness under a fixed token budget.

\begin{figure}
    \centering
    \includegraphics[width=1.0\linewidth]{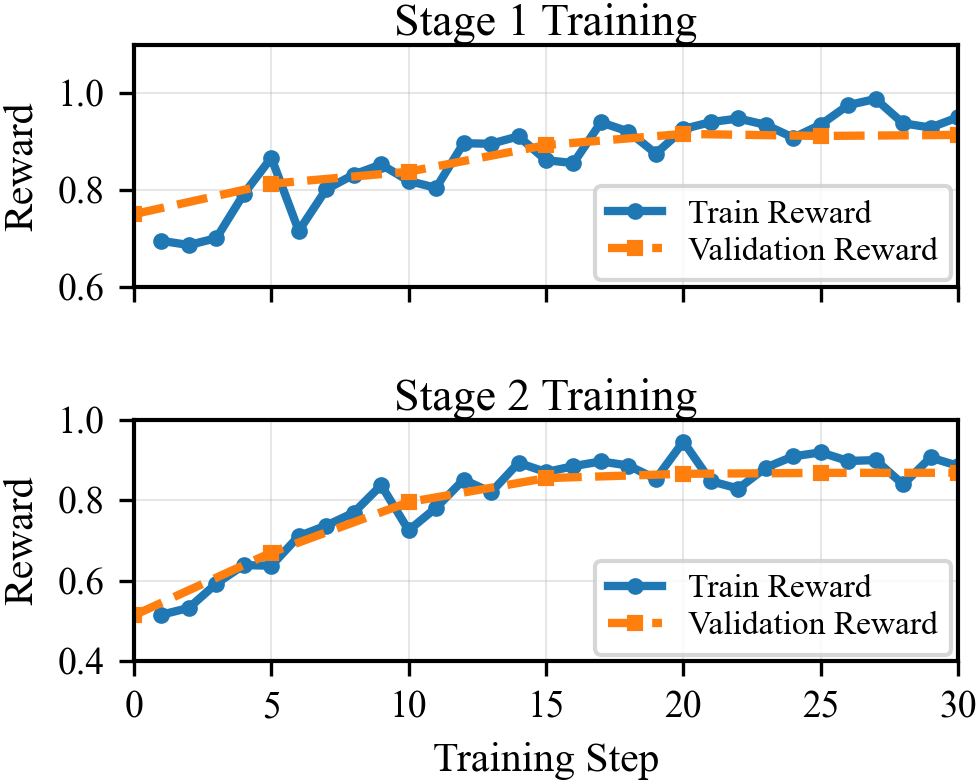}
    \caption{
        \textbf{Training dynamics of Context-Picker using GRPO.} 
        The curves visualize the average reward trajectories on training and validation sets during (Above) the Recall-Oriented Stage I and (Bottom) the Precision-Oriented Stage II. 
        Both stages exhibit stable convergence and a narrow gap between training and validation rewards, indicating that the policy effectively learns to balance evidence coverage and compactness without overfitting.
    }
    \label{fig:curve}
\end{figure}


The reward function is defined as:
\begin{equation}
\small
R_i(o,a)=
\begin{cases}
\text{Cov}(\mathcal{S},\mathcal{S}_{g})-\gamma \dfrac{|\mathcal{S}|-|\mathcal{S}_g|}{L_i}
& \text{if valid and } |\mathcal{S}|\le L_i,\\[6pt]
0
& \text{if valid and } |\mathcal{S}|>L_i,\\
-1.0
& \text{if invalid},
\end{cases}
\label{eq:reward}
\end{equation}
where $\mathcal{S}_g$ is the gold evidence set, $L_i = |\mathcal{S}_g|+\text{red}_i$ is the allowed length with a stage-dependent redundancy margin $\text{red}_i$.

The reward logic follows three principles:
\begin{itemize}
    \item When the output format is valid and the number of selected items does not exceed the ``gold standard + redundancy margin,'' the reward is determined by recall rate with a redundancy penalty proportional to oversampling.
    \item When the selection exceeds the redundancy margin, the reward is set to zero, discouraging excessive evidence inclusion.
    \item When the output format is invalid, a fixed penalty of $-1.0$ is applied to enforce structural correctness.
\end{itemize}

\paragraph{Progressive redundancy compression.}
The key distinction between the two stages lies in the dynamic compression of the redundancy margin \( \text{red} \). Stage~I employs a relaxed margin \( \text{red}_1 \) to tolerate redundancy for completeness, whereas Stage~II tightens the threshold to \( \text{red}_2 \), forcing the policy to eliminate redundant evidence while maintaining high recall. This ``loose-to-tight'' margin adaptation achieves a smooth optimization from \emph{recall sufficiency} to \emph{input compactness}, enabling a Pareto-optimal trade-off between comprehensiveness and efficiency in LCQA.

\paragraph{Stage transition and schedule.}
We implement the two stages as consecutive GRPO phases. We define hyperparameters $T_1$ and $T_2$ to control the number of update steps for the recall-oriented Stage~I and the refinement-oriented Stage~II, respectively. The training is sequential: we first initialize the policy $\pi_\theta$ and train with the Stage~I reward for $T_1$ steps. We then update the reference policy and continue training with the Stage~II reward for $T_2$ steps. In practice, we first train the picker with the Stage~I reward (larger redundancy margin $\text{red}_1$) until the validation reward curve plateaus, and then switch to Stage~II by continuing training from the Stage~I checkpoint with the tighter margin $\text{red}_2$. We found that Context-Picker is robust to the exact split between $T_1$ and $T_2$ as long as Stage~I is given enough updates to learn a high-recall policy; the resulting training dynamics for both stages are shown in Figure~\ref{fig:curve}.

\paragraph{Inference.}
At test time, Context-Picker follows a single-pass retrieve--pick--generate pipeline. Given query $q$ and document $D$, we generate chunks $\mathcal{C}$ and retrieve candidates $\mathcal{C}_{\text{cand}} \leftarrow \text{TopSim}(q, \mathcal{C}; B)$ within budget $B$. We then sample the picker output $\{r, S\} \sim \pi_\theta(\cdot \mid \langle p_{\text{test}}, q, \mathcal{C}_{\text{cand}}\rangle)$, where $S$ is the set of selected IDs. Finally, the generator produces answer $\hat{a} \leftarrow \mathcal{G}(q, \mathcal{C}_{\text{pick}})$ using evidence $\mathcal{C}_{\text{pick}}$ filtered by $S$, which is evaluated against the reference via LLM-as-judge (Section~\ref{sec:experiments}).

\begin{table*}[t]
\centering
\small
\setlength{\tabcolsep}{5pt}
\renewcommand{\arraystretch}{1.15}
\begin{tabular}{llrrrrr}
\toprule
\textbf{Method} & \textbf{Setting} &
\textbf{LoCoMo} & \textbf{MultiFieldQA} & \textbf{HotpotQA} & \textbf{2WikiMQA} & \textbf{MuSiQue} \\
\midrule
\multicolumn{7}{l}{\textit{Standard Baselines}} \\
\multirow{3}{*}{Standard RAG} & TopK=5   &
--    & 0.857 & 0.597 & 0.525 & 0.340 \\
& TopK=10  &
--    & 0.857 & 0.700 & 0.560 & 0.390 \\
& TopK=100 &
0.622 & --    & --    & --    & --    \\
\addlinespace[4pt]

\multicolumn{7}{l}{\textit{Advanced Baselines}} \\
Adaptive-k & Dynamic $k$ &
-- & 0.855 & 0.708 & 0.575 & 0.405 \\
RankRAG & TopK=10 &
0.655 & \textbf{0.875} & 0.732 & 0.665 & 0.465 \\
DynamicRAG & RL-Agent &
0.675 & 0.860 & 0.740 & 0.685 & 0.505 \\
Memory-R1 & RL-Agent &
0.695 & -- & -- & -- & -- \\
\addlinespace[4pt]

\midrule
\multicolumn{7}{l}{\textit{Ours (Context-Picker)}} \\
Context-Picker (Stage 1) & Recall-Oriented &
0.681 & 0.873 & 0.741 & 0.621 & 0.476 \\
Context-Picker (Stage 2) & Precision-Oriented &
\textbf{0.706} & 0.825 & \textbf{0.747} & \textbf{0.702} & \textbf{0.522} \\
\bottomrule
\end{tabular}
\caption{\textbf{Main results on knowledge-intensive QA benchmarks.}
Reported metrics are LLM-as-a-judge Accuracy (Qwen3-32B). Best results are in \textbf{bold}.}
\label{tab:main-results}
\end{table*}

\begin{table}[t]
\centering
\small
\setlength{\tabcolsep}{6pt}
\renewcommand{\arraystretch}{1.15}
\begin{tabular}{lrr}
\toprule
\textbf{Method} & \textbf{Judge Acc (\%)} & \textbf{$\Delta$ vs. full} \\
\midrule
Context-Picker (full)      & 70.6 & --    \\
\quad w/o rationale        & 64.1 & -6.5  \\
\quad w/o redundancy       & 66.0 & -4.6  \\
\quad w/o Stage~1          & 56.5 & -14.1 \\
\bottomrule
\end{tabular}
\caption{\textbf{Ablation study of Context-Picker on \emph{LoCoMo}.}
$\Delta$ denotes absolute drops in Judge Acc (percentage points) compared to the full setting.}
\label{tab:ablation-rlpicker}
\end{table}

\section{Experiments}
\label{sec:experiments}

\subsection{Experimental Setup}

\paragraph{Datasets.}
We evaluate Context-Picker on five knowledge-intensive QA benchmarks that require reasoning over long or multi-hop contexts:
(1) \textbf{LoCoMo} \cite{maharana2024evaluatinglongtermconversationalmemory}, which contains extremely long multi-session conversations and tests long-term conversational memory;
(2) \textbf{MultiFieldQA} \cite{bai2024longbenchbilingualmultitaskbenchmark}, a long-context QA dataset with diverse domains and relatively factoid-style questions;
(3) \textbf{HotpotQA} \cite{yang2018hotpotqadatasetdiverseexplainable}, a classic multi-hop QA benchmark over Wikipedia;
(4) \textbf{2WikiMQA} \cite{ho2020constructingmultihopqadataset}, a multi-hop QA dataset requiring reasoning across two Wikipedia articles;
and (5) \textbf{MuSiQue} \cite{trivedi2022musiquemultihopquestionssinglehop}, which decomposes multi-hop questions into compositional single-hop subquestions.
For datasets in LongBench \cite{bai2024longbenchbilingualmultitaskbenchmark} that do not come with ground-truth evidence passages or IDs, we apply the offline evidence mining procedure in Algorithm~\ref{alg:test-time-evidence-picking} to construct training supervision.
Concretely, we first perform semantic chunking over each long document using \texttt{text-embedding-ada-002} with a similarity threshold of $0.75$, and then mine sufficient and golden evidence sets for each $(q, a)$ pair.

\paragraph{Baselines.}
We compare Context-Picker against three categories of methods: 
(1) \textbf{Standard Baselines}: 
\textbf{Standard RAG} retrieving Top-$K$ ($K \in \{5, 10, 100\}$) chunks.
(2) \textbf{Heuristic/Reranking Methods}: \textbf{Adaptive-$k$}~\cite{taguchi2025efficientcontextselectionlongcontext}, which dynamically selects $k$ via score thresholding, and \textbf{RankRAG}~\cite{NEURIPS2024_db93ccb6} leverages an LLM-based ranker to rerank retrieved passages and select the top-k contexts for generation, thereby improving context ordering.
(3) \textbf{RL-based Agents}: \textbf{DynamicRAG}~\cite{sun2025dynamicragleveragingoutputslarge}, which models the reranker as an RL agent to dynamically adjust the order and number of retrieved documents, and \textbf{Memory-R1}~\cite{yan2025memoryr1enhancinglargelanguage}, which uses RL to learn memory management operations (ADD/UPDATE/DELETE/NOOP).

\paragraph{Evaluation protocol.}
Traditional metrics such as exact match (EM) and F1 are known to be brittle for free-form answers.
For example, the answers ``The cat is on the mat.'' and ``A cat rests on a mat.'' convey essentially the same meaning but would receive a low EM/F1 score due to lexical differences, whereas ``The cat is on the mat.'' and ``The dog is on the mat.'' share substantial n-gram overlap while being factually incompatible.
Following recent work on LLM-based evaluation \cite{gu2025surveyllmasajudge},
we thus adopt an \textbf{LLM-as-judge} protocol as our primary metric.
Given a question $q$ (generated by GPT-4o mini), a reference answer $a^\star$, and a predicted answer $\hat{a}$,
a judge model returns a binary correctness label:
$
\text{Judge}_{\text{ans}}(q, a^\star, \hat{a}) \in \{0, 1\},
$
based on a rubric that checks semantic equivalence to $a^\star$ and penalizes hallucinations or contradictions.
We report the fraction of examples for which the judge predicts correctness, referred to as \emph{Judge Acc}.

\subsection{Main Results}
\label{subsec:main-results}

Table~\ref{tab:main-results} summarizes the main results across the five benchmarks.

\paragraph{Performance comparison.}
Table~\ref{tab:main-results} presents the comparative results. Context-Picker (Stage 2) achieves the best performance on four out of five benchmarks, outperforming both standard RAG and advanced baselines.
Specifically, on reasoning-heavy datasets like \textbf{MuSiQue} and \textbf{HotpotQA}, our method surpasses the strong reranking baseline RankRAG by margins of $+5.7\%$ and $+1.5\%$, respectively. This highlights that for multi-hop questions, simply reordering documents (RankRAG) is insufficient; explicitly pruning distractors to form a minimal evidence set is crucial for accurate reasoning.
Compared to RL-based peers (DynamicRAG, Memory-R1), Context-Picker also demonstrates superior accuracy, particularly on the ultra-long \textbf{LoCoMo} benchmark.

\paragraph{Recall vs. Precision Trade-off.}
On \textbf{MultiFieldQA}, RankRAG achieves the top score, slightly outperforming our Precision-Oriented Stage~II.
We attribute this to the dataset's \emph{needle-in-a-haystack} nature: many questions are closer to extractive factoid QA, where answerablity is dominated by whether a small number of critical snippets are retrieved, and the benefit of redundancy suppression is comparatively limited.
Our Recall-Oriented Stage~I reaches $0.873$, statistically tied with RankRAG, indicating that the coarse stage effectively maximizes evidence coverage.
In contrast, Stage~II enforces a tighter redundancy margin and explicitly optimizes evidence compression; while this is advantageous on reasoning-heavy benchmarks by removing distractors (e.g., \textbf{MuSiQue}), it can occasionally prune borderline but helpful cues, leading to a slight drop on MultiFieldQA.
Overall, the two-stage design provides a practical controllable trade-off: Stage~I prioritizes recall, while Stage~II prioritizes compactness and reasoning density.

\paragraph{Effect of the two-stage schedule.}
The two-stage training scheme yields a clear pattern.
Stage~I, which uses a relaxed redundancy margin and emphasizes high recall, is particularly helpful on datasets where evidence is dispersed or conversations are long.
Stage~II, which tightens the redundancy penalty to favor minimal sufficient sets, further improves accuracy on four out of five benchmarks while also shortening the selected contexts.
This supports our hypothesis that gradually shifting the objective from recall to precision leads to a better quality–efficiency trade-off than optimizing a single-stage objective.

\paragraph{Training stability.}
Reinforcement learning on discrete text selection is often characterized by instability. 
However, thanks to our dense reward supervision mined via LOO and the GRPO algorithm, Context-Picker demonstrates robust training dynamics. 
As illustrated in Figure~\ref{fig:curve}, the reward curves for both the Recall-Oriented Stage~I (Above) and Precision-Oriented Stage~II (Bottom) show steady convergence.
The minimal gap between training and validation performance further validates the generalization capability of our offline evidence mining strategy.

\subsection{Ablation Studies}
\label{subsec:ablation}

To better understand which components of Context-Picker drive the gains, we conduct ablations on the \emph{LoCoMo} dataset.
We focus on three aspects: rationale generation, redundancy-aware reward shaping, and the recall-oriented Stage~I.

\paragraph{Rationale generation.}
In the full model, the picker outputs both a short natural-language rationale and a set of selected IDs.
Removing the rationale branch (``w/o rationale'') leads to a $6.5$-point drop in Judge Acc.
We hypothesize that requiring the model to verbalize why certain passages are selected acts as a structural regularizer:
it encourages more stable reasoning over evidence interactions and reduces the tendency to over-select loosely related passages.

\paragraph{Redundancy-aware reward shaping.}
When we remove the redundancy term in the reward (``w/o redundancy''),
the picker no longer receives explicit penalties for overshooting the golden set size.
Under the same token budget, this variant tends to keep more passages and accumulates noise,
resulting in a $4.6$-point drop on LoCoMo.
This confirms that explicitly modeling length/redundancy in the reward is important for achieving a good balance between recall and precision, rather than relying solely on an implicit budget constraint.

\paragraph{Role of the recall-oriented Stage~I.}
Finally, we examine a variant trained only with the Stage~II objective (``w/o Stage~1''),
i.e., directly optimizing the refinement-oriented reward from scratch.
This leads to a substantial degradation to $56.5\%$ Judge Acc, $14.1$ points below the full two-stage Context-Picker.
Although the ``w/o Stage 1" variant optimizes the same final reward function, it suffers from the cold-start problem inherent in combinatorial optimization. The Stage 2 reward imposes a strict penalty on redundancy. Without the high-recall initialization provided by Stage 1, the agent faces a sparse reward landscape: exploring larger subsets incurs immediate length penalties, while small random subsets rarely contain the complete evidence chain required for reasoning. Consequently, the policy gets trapped in a local optimum of selecting overly sparse contexts to minimize penalties, failing to answer the question.

These ablation studies show that our coarse-to-fine optimization schedule, the redundancy-aware reward shaping, along with the rationale generated by the policy, all contribute to our method.

\section{Related Work}

\paragraph{Adaptive RAG.} 
While standard RAG retrieves a fixed top-$K$ passages~\cite{lewis2021retrievalaugmentedgenerationknowledgeintensivenlp}, recent works attempt to adapt context dynamically. Approaches like Self-RAG~\cite{asai2023selfraglearningretrievegenerate} and FLARE~\cite{jiang2023activeretrievalaugmentedgeneration} interleave retrieval with generation, but incur high inference latency. Adaptive-RAG~\cite{jeong2024adaptiveraglearningadaptretrievalaugmented} and Adaptive-$k$~\cite{taguchi2025efficientcontextselectionlongcontext} select retrieval depth via classifiers or score thresholds, yet they rely on heuristics rather than optimizing for evidence sufficiency. Unlike rerankers that score passages individually~\cite{sun2024chatgptgoodsearchinvestigating,wang2024learningretrieveincontextexamples}, Context-Picker optimizes the \emph{evidence subset} directly.

\paragraph{RL for Retrieval.} 
RL has been applied to query reformulation~\cite{zhu2025convsearchr1enhancingqueryreformulation} and retriever alignment~\cite{sun2025dynamicragleveragingoutputslarge}. However, methods like DynamicRAG typically optimize for ranking quality or answer correctness via sparse, trajectory-level rewards. Recent memory-agents~\cite{yan2025memoryr1enhancinglargelanguage,maharana2024evaluatinglongtermconversationalmemory} use RL for long-term memory management but often lack mechanisms to penalize redundancy explicitly. Our framework differs by using offline-mined ``minimal sufficient sets" to provide dense supervision, enabling the policy to learn precise evidence compression.

\section{Conclusion}
We introduced \emph{Context-Picker}, a reinforcement learning framework that shifts context retrieval from static ranking to dynamic, minimal sufficient subset selection. By leveraging offline-mined supervision and a two-stage coarse-to-fine training schedule, our model learns to balance high recall with strict compactness. Experiments across five benchmarks demonstrate that Context-Picker outperforms strong RAG baselines in answer accuracy. Future work will extend this paradigm to open-ended generation tasks and investigate integration with token-level compression techniques to further optimize inference costs.


\bibliographystyle{named}
\bibliography{ijcai26}

\end{document}